\documentclass[11pt,a4paper]{article}
\usepackage[hyperref]{emnlp2020}
\usepackage{times}
\usepackage{latexsym}

\usepackage{microtype}

\aclfinalcopy 
\def\aclpaperid{2835} 

\usepackage{amsmath}
\usepackage{amssymb}
\usepackage{hyperref}
\usepackage{booktabs}
\usepackage{pgfplots}
\pgfplotsset{width=.9\columnwidth}
\usepackage{multirow}
\usepackage{caption}
\usepackage{subcaption}
\usepackage{tikz}
\usepackage{xspace}
\usepackage{xargs}
\usepackage[colorinlistoftodos,prependcaption,textsize=tiny]{todonotes}
\usepackage{multirow}
\usetikzlibrary{positioning}
\usepackage{pifont}

\usepackage{makecell}
\usepackage{verbatim}

\newcommand\ignore[1]{}

\def\model/{\textsc{ReConsider}}

\newcommandx{\sy}[2][1=]{\todo[linecolor=purple,backgroundcolor=purple!10,bordercolor=purple,#1]{Scott: #2}\xspace}
\newcommandx{\dc}[2][1=]{\todo[linecolor=blue,backgroundcolor=blue!10,bordercolor=blue,#1]{Danqi: #2}\xspace}

\newcommand\tf[1]{\textbf{#1}}

\pgfplotsset{compat=1.14}

\title{\model/: Re-Ranking using Span-Focused Cross-Attention for Open Domain Question Answering}
\date{}

\author{
    Srinivasan Iyer$^\dagger$\quad Sewon Min$^\ddagger \thanks{\hspace{.06in}Work done while at Facebook AI.}$\quad Yashar Mehdad$^\dagger$\quad Wen-tau Yih$^\dagger$ \\
    $^\dagger$Facebook AI \qquad $^\ddagger$University of Washington \\
    \texttt{\{sviyer,mehdad,scottyih\}@fb.com} \qquad \texttt{sewon@cs.washington.edu} \\ 
 }

\begin{document}
\maketitle

\begin{abstract}
State-of-the-art Machine Reading Comprehension (MRC) models for Open-domain Question Answering (QA) are typically trained for span selection using distantly supervised positive examples and heuristically retrieved negative examples. 
This training scheme possibly explains empirical observations that these models achieve a high recall amongst their top few predictions, but a low overall accuracy, motivating the need for answer re-ranking. 
We develop a simple and effective re-ranking approach (\model/) for span-extraction tasks, that improves upon the performance of large pre-trained MRC models. \model/ is trained on positive and negative examples extracted from high confidence predictions of MRC models, and uses in-passage span annotations to perform span-focused re-ranking over a smaller candidate set. As a result, \model/ learns to eliminate close false positive passages, and achieves a new state of the art on four QA tasks, including 45.5\% Exact Match accuracy on Natural Questions with real user questions, and 61.7\% on TriviaQA.
\end{abstract}

\newcommand{\triviaqa}{\textsc{TriviaQA}}
\newcommand{\nq}{\textsc{NQ}}
\newcommand{\webq}{\textsc{WebQ}}
\newcommand{\squad}{\textsc{SQuAD}}
\newcommand{\trec}{\textsc{TREC}}

\newcommand{\sota}{state-of-the-art}
\newcommand{\bert}{\textsc{BERT}}
\newcommand{\roberta}{\textsc{RoBERTa}}
\newcommand{\bertbase}{\textsc{BERT}$_\mathrm{base}$}
\newcommand{\bertlarge}{\textsc{BERT}$_\mathrm{large}$}
\newcommand{\bartlarge}{\textsc{BART}$_\mathrm{large}$}
\newcommand{\robertalarge}{\textsc{RoBERTa}$_\mathrm{large}$}
\newcommand{\dpr}{\textsc{DPR}}

\section{Introduction}

Open-domain Question Answering \cite{voorhees1999trec} (QA) involves answering questions by extracting correct answer spans from a large corpus of passages, and is typically accomplished by a  light-weight passage retrieval model followed by a heavier Machine Reading Comprehension (MRC) model~\citep{chen2017reading}. The span selection components of MRC models are trained on distantly supervised positive examples (containing the answer string) together with heuristically chosen negative examples, typically from upstream retrieval models. This training scheme possibly explains empirical findings \cite{wang2017evidence,wang2018multi} that while MRC models can confidently identify top-K answer candidates (high recall), they cannot effectively discriminate between top semantically similar false positive candidates (low accuracy). In this paper, we propose a simple and generally applicable span-focused re-ranking approach for extractive tasks, that is trained on the high confidence predictions of MRC models and achieves state of the art results on four open-domain QA datasets.

\begin{figure}
    \centering
    \resizebox{\columnwidth}{!}{
    \includegraphics[width=\textwidth]{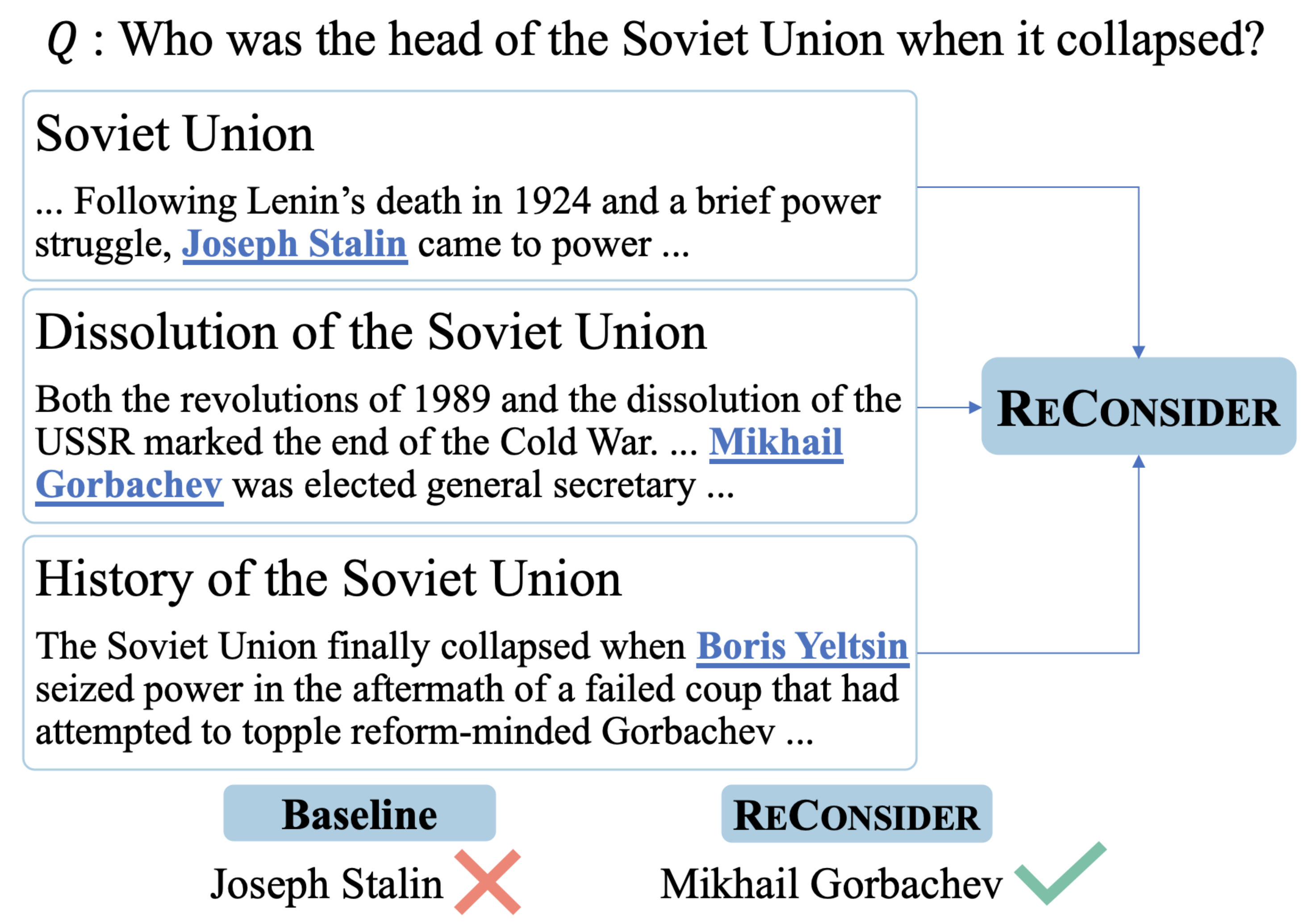}}
    \caption{
        Top-3 passage-spans predicted by a BERT-MRC model on a question from NQ (answer spans are underlined). \model/ re-evaluates the passages with marked candidate answers, eliminates close false positives and ranks \textit{Mikhail Gorbachev} as correct.
    }
    \label{fig:intro}
\end{figure}

Earlier work \cite{wang2018multi,wang2017evidence} on open-domain QA have recognized the potential of answer re-ranking, and we observe that this potential persists despite recent advances using large pre-trained models like \bert ~\cite{devlin2018bert}. Figure \ref{fig:intro} shows the top-3 predictions of a \bert-based SOTA model \cite{karpukhin2020dense} on a question from the Natural Questions (NQ)~\citep{kwiatkowski2019natural} dataset, ``\textit{Who was the head of the Soviet Union when it collapsed?}" While all predictions are highly relevant and refer to heads of the Soviet Union, \textit{Mikhail Gorbachev} is correct and the rest are close false positives. Table \ref{tab:oracle} presents accuracies obtained by the same model on four QA datasets, if the answer exactly matches any of the top-K predictions for $k=1,5,10$ and $25$. We observe that an additional $10\%$ and $~20\%$ of correct answers exist amongst the top-5 and top-25 candidates respectively, presenting an enormous opportunity for span reranking models.

\begin{table}[t]
    \setlength\tabcolsep{5pt}
    \centering
    \begin{tabular}{lcccc} 
    \toprule
    {Dataset} & {Top-1} & {Top-5} & {Top-10} & {Top-25} \\ 
    \midrule
    \nq & 40.3 & 49.5 & 50.9 & 62.4  \\
    \triviaqa & 57.2 & 64.6 & 65.7 & 73.1 \\ 
    \webq & 42.6 & 49.0 & 50.7 & 60.4 \\
    \trec & 49.6 & 58.7 & 60.9 & 71.4 \\
    \bottomrule
    \end{tabular}
    \caption{Top-k EM accuracies using a \sota\ model~\citep{karpukhin2020dense} on four open-domain QA tasks (dev set). Improvements of up to 22\% can potentially be achieved by re-ranking top-25 candidates.}
    \label{tab:oracle}
\end{table}

Our re-ranking model is trained using positive and negative examples extracted from high confidence MRC model predictions, and thus, learns to eliminate hard false positives. This can be viewed as a coarse-to-fine approach of training span selectors, with the base MRC model trained on heuristically chosen negatives and the re-ranker trained on finer, more subtle negatives. This contrasts with multi-task training approaches \cite{wang2018multi}, whose re-scoring gains are limited by training on the same data, especially when coupled with large pre-trained models. Our approach also scales to any number of ranked candidates, unlike previous concatenation based cross-passage re-ranking methods \cite{wang2017evidence} that do not transfer well to current length-bounded large pre-trained models. 
Similar to MRC models, our re-ranking approach uses cross-attention between the question and a candidate passage \cite{seo2016bidirectional}. However, we now mark each passage with a specific candidate answer span, to assist the model to perform focused reasoning relative to the annotated span, in contrast to MRC models, which must reason across all spans. Therefore, the re-ranker performs span ranking of carefully chosen candidates, rather than span selection like the MRC model. Similar focused cross-attention methods have recently proved to be effective for Entity Linking \cite{wu2019zero} tasks, although they annotate the inputs instead of the retrieved passages. 

We use our span-focused approach to re-rank MRC model predictions from \newcite{karpukhin2020dense} and achieve a new extractive \sota~on four QA datasets, including 45.5\% on NQ (real user queries, +1.6\% on small models) and 61.1\% on TriviaQA~\citep{joshi-etal-2017-triviaqa} (+2.5\% on small models). Our approach is easy to implement and broadly applicable to other span-extractive tasks.

\section{Background}
\label{sec:background}

Open-domain Question Answering (QA) aims to answer factoid questions from a large corpus of passages~\citep{voorhees1999trec} (such as Wikipedia) in contrast with single passage MRC tasks \cite{rajpurkar2016squad}. Prior works use pipelined approaches, that first retrieve candidate passages and subsequently use a neural MRC model to extract answer spans~\citep{chen2017reading}, with further improvements using joint learning \cite{wang2018r,tan2017s}. Recent successes involve improving  retrieval, thereby increasing the coverage of passages fed into the MRC model~\citep{guu2020realm,karpukhin2020dense}. In this paper, we significantly improve upon the relatively understudied MRC models by performing a simple span-focused re-ranking of its highly confident predictions.

For Open-domain QA, it is crucial to train MRC models to distinguish passage-span pairs containing the answer ({\em positives}) from those that do not ({\em negatives}). Using negatives that appear as close false positives can produce more robust MRC models. However, prior work relies on upstream retrieval models to supply distantly supervised positives (contain answer string) and negatives ~\citep{asai2020learning}, that are in-turn trained using heuristically chosen positives and negatives. Our approach leverages positives and negatives from highly confident MRC predictions which are hard to classify, and thus, improve upon MRC model performance.

\newcite{jia2017adversarial} motivate recent work on answer verification for QA by showing that MRC models are easily confused by similar passages.
\newcite{wang2017evidence} use a weighted combination of three re-rankers and rescore a concatenation of all passages with a particular answer using a sequential model, while, \newcite{wang2018multi} develop a multi-task end-to-end answer scoring approach. Although the main idea is to consider multiple passage-span candidates collectively, such approaches either used concatenation, which is prohibitively expensive to couple with length-restricted models like \bert, or are trained on the same data without variations only to realize marginal gains. \newcite{hu2019read+} use answer verification to predict the unanswerability of a question-passage pair for traditional MRC tasks. To our knowledge, our work is the first to (i) propose a simple re-ranking approach that significantly improves over large pre-trained models~\citep{devlin2018bert} in an open domain setting, and (ii) use annotated top model predictions as harder negatives to train more robust models for QA.

\section{Model}
\label{sec:model}

We assume an extractive MRC model $\mathcal{M}$ coupled with a passage retrieval model, that given a question $q$ and a passage corpus
$\mathcal{P}$, produces
a list of $N$ passage and span pairs, $\{(p_j, s_j)\}_{j=1}^{N}, p_j \in \mathcal{P}$ and $s_j$ is a span within $p_j$, ranked by the likelihood of $s_j$ answering $q$.
Note that
$\{p_j\}_{j=1}^N$ is not disjoint as more than one span can come from the same passage.
In this section, we develop a simple span-focused re-ranking model $\mathcal{R}$, that learns a distribution $p$, over top-$K$ $(p_j, s_j)$ pairs $1\leq j \leq K$, given question $q$. 
Essentially, model $\mathcal{R}$ first scores every $(q, p_j, s_j)$ triple using scoring function $r$, and then normalizes over these scores to produce $p$:
\begin{align}
    p(q, p_j, s_j) = \frac{e^{r(q, p_j, s_j)}}{\sum_{1 \leq k \leq K} e^{r(q, p_k, s_k)}}.
\end{align}

Specifically, if $E(q, p_j, s_j) \in \mathbb{R}^H$ is a dense representation of $(q, p_j, s_j)$, $r$ is defined as:
\begin{align}
    r(q, p_j, s_j) &= \mathbf{w}^T E(q, p_j, s_j),
\end{align}
where $\mathbf{w} \in \mathbb{R}^H$ is a learnable vector.

\paragraph{Span-focused tuple encoding} We compute $E$ using the representation of the \texttt{[CLS]} token of a BERT model \cite{devlin2018bert} applied to a span-focused encoding of $(q, p_j, s_j)$. This encoding is generated by first marking the tokens of $s_j$ within passage $p_j$ with special start and end symbols \texttt{[A]} and \texttt{[/A]}, to form $\hat{p}_j$, followed by concatenating the \texttt{[CLS]} and question tokens, with the annotated passage tokens $\hat{p}_j$, using separator token \texttt{[SEP]}. We find span marking to be a crucial ingredient for answer re-ranking, without which, performance deteriorates (Section \ref{sec:exp}).

\paragraph{Training} We obtain top $K$ predictions $(p_j, s_j)$ of model $\mathcal{M}$ for each question $q_i$ in its training set, which we divide into positives, where $s_j$ is exactly the groundtruth answer, and remaining negatives. We train $\mathcal{R}$ using mini-batch gradient descent, where in each iteration, for question $q$, we include 1 randomly chosen positive and $M - 1$ randomly chosen negatives, and maximize the likelihood of the positive. Unlike the heuristically chosen negatives used to train $\mathcal{M}$, $\mathcal{R}$ is trained using negatives from high confidence predictions of $\mathcal{M}$, which are harder to classify. Thus, this can be viewed as an effective coarse-to-fine negative selection strategy for span extraction models (Section~\ref{sec:exp}). 

\section{Baseline Model $\mathcal{M}$}
\label{sec:baseline}

We use the state-of-the-art models of \newcite{karpukhin2020dense} which consists of 1) a dense passage retriever, and 2) a span extractive \bert~reader, as our model $\mathcal{M}$. 
The retriever uses a passage encoder $f_p$ and a question encoder $f_q$ to represent all passages and questions as dense vectors in the same space. 
During inference, it retrieves top-100 passages similar to question $q$ based on their inner product, which are subsequently passed on to the span-extractive MRC reader. 

\newcommand{\nqbm}{32.6}
\newcommand{\nqsingle}{41.5}
\newcommand{\nqsinglehybrid}{39.0}
\newcommand{\nqmulti}{41.5}
\newcommand{\nqmultihybrid}{38.8}

\newcommand{\triviabm}{52.4}
\newcommand{\triviasingle}{56.8}
\newcommand{\triviasinglehybrid}{57.0}
\newcommand{\triviamulti}{56.8}
\newcommand{\triviamultihybrid}{57.9}

\newcommand{\sqbm}{38.1}
\newcommand{\sqsingle}{29.8}
\newcommand{\sqsinglehybrid}{36.7}
\newcommand{\sqmulti}{24.1}
\newcommand{\sqmultihybrid}{35.8}

\newcommand{\webqbm}{29.9}
\newcommand{\webqsingle}{34.6}
\newcommand{\webqsinglehybrid}{35.2}
\newcommand{\webqmulti}{42.4}
\newcommand{\webqmultihybrid}{41.1}

\newcommand{\trecbm}{24.9}
\newcommand{\trecsingle}{25.9}
\newcommand{\trecsinglehybrid}{28.0}
\newcommand{\trecmulti}{49.4}
\newcommand{\trecmultihybrid}{50.6}

\begin{table*}[t]
    \setlength\tabcolsep{5pt}
    \centering
    \begin{tabular}{lccccc} \toprule
    {Model} & {\nq} & {\triviaqa} & {\webq} & {\trec}  \\ \midrule
    BM25+BERT~\cite{lee2019latent} & 26.5 & 47.1 & 17.7 & 21.3  \\
    {ORQA}~\cite{lee2019latent} & 33.3 & 45.0 & 36.4 & 30.1    \\
    HardEM~\cite{min2019discrete} & 28.1 & 50.9 & - & -  \\
    GraphRetriever~\cite{min2019knowledge} & 34.5 & 56.0 & 36.4 & -  \\
    PathRetriever~\cite{asai2020learning}  & 32.6 & - & - & - \\
    {REALM}~\cite{guu2020realm} & 39.2 & - & 40.2 & 46.8  \\ 
    {REALM}$_\textrm{News}$~\cite{guu2020realm} & 40.4 & - & 40.7 & 42.9  \\ 
    \midrule
        \tf{Models that use \dpr$_\textrm{multi}$} \cite{karpukhin2020dense} \\
    \midrule
    DPR-BERT$_\textrm{base}$~\cite{karpukhin2020dense} & \nqmulti & \triviamulti & \webqmulti & \trecmulti  \\
    \model/$_\textrm{base}$ (Ours)& \tf{43.1} & \tf{59.3} & \tf{44.4}  & \tf{49.3}  \\
    \midrule
    DPR-BERT$_\textrm{large}^\dagger$~\cite{karpukhin2020dense}& 44.6 & 60.9 & 44.8 & 53.5 \\
    RAG$_\textrm{large}$~\cite{lewis2020retrievalaugmented} & 44.5 & 56.1 & 45.5 & 52.2 \\
    \model/$_\textrm{large}$ (Ours)& \tf{45.5} & \tf{61.7} & \tf{45.9} & \tf{55.3} \\
    \bottomrule
    \end{tabular}
    \caption{
    End-to-end QA (Exact Match) Accuracy.  Models in the lower half use dense passage retrieval from~\newcite{karpukhin2020dense}. \model/ outperforms previous methods under both base and large versions. {\footnotesize Dataset statistics and dev set results in Appendix. `$\textrm{News}$': news articles are added;
    $^\dagger$: numbers from our own experiments.}
    }
    \label{tab:qa_em}
\end{table*}

The MRC reader is an extension of model $\mathcal{R}$ of Section \ref{sec:model}, to perform span extraction. We briefly describe it but \newcite{karpukhin2020dense} has complete details. Its input is a question $q$ together with positive and negative passages $p_j$ from its retrieval model. $(q, p_j)$ tuples are encoded as before ($ enc(q, p_j) = q$ \texttt{[SEP]} $p_j$), but without spans being marked (as spans are unavailable). A distribution over passages $p_s$ is computed as before using scoring function $r$ and context encoder $E$. In addition, a start-span probability, $p_{st}(t_i|q,p_j)$ and an end-span probability, $p_{e}(t_i|q,p_j)$ is computed for every token $t_i$ in $enc(q, p_j)$. The model is trained to maximize the likelihood of $p_s(p_j) \times p_{st}(s|q,p_j) \times p_{e}(t|q,p_j)$ for each correct answer span $(s, t)$ in $p_j$, and outputs the top-K scoring passage-span pairs during inference.


\section{Experiments}
\label{sec:exp}

\paragraph{Datasets} We use four benchmark open-domain QA datasets following \newcite{lee2019latent}: \\
\tf{Natural Questions (\nq)} contains real user questions asked on Google searches; we consider questions with short answers up to 5 tokens. \\
\tf{\triviaqa} \cite{joshi-etal-2017-triviaqa} consists of questions collected from trivia and quiz-league websites; we take questions in an unfiltered setting and discard the provided web snippets.\\
\tf{WebQuestions (\webq)} \cite{berant2013semantic} is a collection of questions extracted from the Google Suggest API, with answers being Freebase entities. \\
\tf{CuratedTREC} \cite{baudivs2015modeling} contains curated questions from TREC QA track.

\paragraph{Implementation details} For all datasets, we use the retrieval model (without retraining) and setup from~\newcite{karpukhin2020dense}, retrieving 100-token passages from a Wikipedia corpus (from $2018$-$12$-$20$). We also use their MRC model with their best performing hyperparameters as model $\mathcal{M}$. For model $\mathcal{R}$, we experiment with both \bertbase\ and \bertlarge, use top-100 predictions from model $\mathcal{M}$ during training (top-5 for testing), and use $M=30$ for \bertbase\ and $M=10$ for \bertlarge. We use a batch size of $16$ on \nq\ and \triviaqa~and $4$ otherwise. For \webq\ and \trec, we start training from our trained \nq~model. We will release all related data, models and code.

\paragraph{Results} We present end-to-end exact match accuracy results on the test set of these datasets in Table \ref{tab:qa_em} compared with previous models. The \bertbase~version of \model/ outperforms the previous \sota~DPR model of \newcite{karpukhin2020dense} (our model $\mathcal{M}$) by $1.6\%$ on \nq~and $\sim2\%$ on \triviaqa~and \webq. For training on the smaller datasets, \webq~and \trec, we initialize models using the corresponding \nq~model. Table \ref{tab:qa_em} demonstrates the effectiveness of a coarse-to-fine approach for selecting negative passages, with dense retrieval based negatives (DPR) outperforming BM25, and in-turn, improved upon using our span-focused re-ranking approach. We obtain improvements despite $\mathcal{R}$ being not only very similar in architecture to the MRC reader $\mathcal{M}$, but also trained on the same QA pairs, owing to (i) training using harder false-positive style negatives, and (ii) answer-span annotations that facilitate a re-allocation of modeling capacity from modeling all spans to reasoning about specific spans with respect to the question and the passage. Re-ranking performance suffers without the use of these methods. For example, substituting answer-span annotations with answer concatenation reduces accuracy by ~1\% on the validation set of \nq.

We train a large variant of \model/ using \bertlarge~for model $\mathcal{R}$, trained on predictions from a \bertlarge~model $\mathcal{M}$.
For a fair comparison, we re-evaluate DPR using \bertlarge. \model/$_\textrm{large}$ outperforms it by $\sim1\%$ on all datasets (+ $\sim2\%$ on \trec). This model is also comparable in size to RAG \cite{lewis2020retrievalaugmented} (which uses \bartlarge) but outperforms it on all tasks (+1 on \nq, +5.5 on \triviaqa, +3 on \trec), demonstrating that retrieve-extract architectures can match the performance of answer generation models.

We find $K$=5 (testing) to be best for all datasets, and increasing $K$ has little effect on accuracy, despite training the model on top-100 predictions. Although in contrast with our expectations based on Table \ref{tab:oracle}, this is anticipated since very low-ranked predictions are less likely to be reranked highly, but this also presents an opportunity for future work.

\section{Conclusion}
\label{sec:conclusion}

We present a simple and effective re-ranking approach for span-extractive models, that works using a synergistic combination of two techniques viz. retraining with harder negatives, and, span-focused cross attention. This method achieves SOTA results on four open domain QA datasets, also outperforming recent generative pre-training approaches.

\bibliography{emnlp2020}
\bibliographystyle{acl_natbib}

\clearpage

\appendix
\section{Computing Infrastructure Used}

All experiments were run on a machine with 2 chips of Intel(R) Xeon(R) CPU E5-2698 v4 @ 2.20GHz with 20 cores (40 threads) each, equipped with $8$ NVIDIA TESLA V100 GPUs, each with $32$ GB of memory.  

\section{Average Run-time and \#Parameters}

We report average run-times for training and inference on \nq~(\triviaqa~is similar), as well as number of model parameters, in Table \ref{tab:runtime}. \webq~and~\trec~are much smaller datasets and have lower runtimes.

\begin{table}[h]
    \centering
    \begin{tabular}{lccc} 
    \toprule
    \tf{Model} & \tf{Params} & \tf{Train} & \tf{Inf.} \\ 
    \midrule
    DPR-BERT$_\textrm{large}$ & 335M & 37 h & 2.8 h \\
    \model/$_\textrm{base}$ & 109M & 13 h & 2 m \\
    \model/$_\textrm{large}$ & 335M & 28 h & 2 m  \\
    \bottomrule
    \end{tabular}
    \caption{Runtime for training and inference, and number of parameters of the models that we executed, on \nq. Runtimes for \model/ do not include the time required to train and obtain predictions from DPR. }
    \label{tab:runtime}
\end{table}

\begin{table}
    \setlength\tabcolsep{5pt}
    \centering
    \begin{tabular}{lccccc} \toprule
    \tf{Model} & \tf{\nq} & \tf{\triviaqa} & \tf{\webq} & \tf{\trec}  \\ \midrule
    \model/$_\textrm{base}$ (Ours)& 42.5 & 60 & 46.3 & 49.6  \\
    \midrule
    DPR-BERT$_\textrm{large}$~\cite{karpukhin2020dense}& 42.2 & 60.1 & 43.8 & 54.9  \\
    \model/$_\textrm{large}$ (Ours)& 44.2 & 61.7 & 46 & 54.9 \\
    \bottomrule
    \end{tabular}
    \caption{Validation set performance for our experiments.
    }
    \label{tab:dev}
\end{table}

\section{Hyperparameters}

For training \model/, we use top-$100$ predictions of the baseline MRC model. This was chosen from $\{50, 75, 100\}$ based on validation set EM.  For training \model/, we use 1 positive and $M - 1$ negatives during each iteration. We tried values of $M$ between $5$ and $40$ in increments of $5$ and chose $M=30$ based on validation set EM ($M=10$ for \bertlarge). Similarly, we re-rank $K=5$ candidates during inference, and this value was chosen by experimenting with values $2,3,4$ and values between $5$ and $20$, in increments of $5$.

\section{Validation Performance}

Table \ref{tab:dev} presents validation set performance for the experiments that we ran for this paper.

\section{Dataset Statistics}

Table \ref{tab:data} presents the number of examples in the training, validation and testing splits of the four open-domain QA datasets that we use, based on the dataset prepared by \newcite{karpukhin2020dense}. 

\begin{table}
    \centering
    \begin{tabular}{lrrrr} \toprule
    \tf{Dataset} & \tf{Train}  & \tf{Dev} & \tf{Test} \\
    \midrule
    \nq   &     67,098 & 8,757 & 3,610 \\
    \triviaqa &  67,975 & 8,837 & 11,313  \\
    \webq    &   2,898 & 361   & 2,032 \\
    \trec   &    1,240 & 133   & 694 \\
    \bottomrule
    \end{tabular}
    \caption{Training, validation and testing set sizes for the four open-domain QA tasks evaluated in our paper.}
    \label{tab:data}
\end{table}

\end{document}